\documentclass[journal]{IEEEtran}
\usepackage{times}
\usepackage{amssymb}
\usepackage{helvet}
\usepackage{mathrsfs}
\usepackage{url}
\usepackage{courier}
\usepackage{booktabs,caption}
\usepackage{threeparttable}
\usepackage{amsmath}
\usepackage{graphicx}
\usepackage{booktabs}
\usepackage{url}	
\usepackage{subfigure}
\usepackage{overpic}
\usepackage{color}
\usepackage[english]{babel}
\usepackage[T1]{fontenc}
\usepackage[utf8]{inputenc}
\usepackage{multirow}
\usepackage{diagbox}
\usepackage{authblk}
\usepackage[numbers, compress]{natbib}
\begin{document}
\title{H-DenseUNet: Hybrid Densely Connected UNet for Liver and Tumor Segmentation from CT Volumes}
%
\author{Xiaomeng Li, Hao Chen, \emph{Member, IEEE}, Xiaojuan Qi, Qi Dou, \emph{Student Member, IEEE}, \\
Chi-Wing Fu, \emph{Member, IEEE} and Pheng-Ann Heng, \emph{Senior Member, IEEE} \thanks{This work was supported by the grants from the Research Grants Council of the Hong Kong Special Administrative Region (Project Nos.  GRF 14202514 and GRF 14203115).} \thanks{
X. M. Li, H. Chen, X. J. Qi, Q. Dou, C. W. Fu and P. A. Heng are with the Department of Computer Science and Engineering, The Chinese University of Hong Kong, Hong Kong. H.Chen is also with Imsight Medical Technology, Inc. 
(e-mail: xmli@cse.cuhk.edu.hk; hchen@cse.cuhk.edu.hk). Corresponding author: Hao Chen}}

%

\markboth{IEEE Transactions on Medical Imaging}%
{Shell \MakeLowercase{\textit{et al.}}: Bare Demo of IEEEtran.cls for IEEE Journals}
%



\newcommand{\revise}[1]{{\color{black}{#1}}}

\maketitle

\begin{abstract}
Liver cancer is one of the leading causes of cancer death. To assist doctors in hepatocellular carcinoma diagnosis and treatment planning, an accurate and automatic liver and tumor segmentation method is highly demanded in the clinical practice.
Recently, fully convolutional neural networks (FCNs), including 2D and 3D FCNs, serve as the back-bone in many volumetric image segmentation. However, 2D convolutions can not fully leverage the spatial information along the third dimension while 3D convolutions suffer from high computational cost and GPU memory consumption.
To address these issues, we propose a novel hybrid densely connected UNet (H-DenseUNet), which consists of a 2D DenseUNet for efficiently extracting intra-slice features and a 3D counterpart for hierarchically aggregating volumetric contexts under the spirit of the auto-context algorithm for liver and tumor segmentation.
We formulate the learning process of H-DenseUNet in an end-to-end manner, where the intra-slice representations and inter-slice features can be jointly optimized through a hybrid feature fusion (HFF) layer.
We extensively evaluated our method on the dataset of MICCAI 2017 Liver Tumor Segmentation (LiTS) Challenge and 3DIRCADb Dataset.
Our method outperformed other state-of-the-arts on the segmentation results of tumors and achieved very competitive performance for liver segmentation even with a single model. 
\end{abstract}

\IEEEpeerreviewmaketitle

\section{Introduction}
Liver cancer is one of the most common cancer diseases in the world and causes massive deaths every year~\citep{ferlay2010estimates,lu2006liver}.
The accurate measurements from CT, including tumor volume, shape, location and further functional liver volume, can assist doctors in making accurate hepatocellular carcinoma evaluation and treatment planning. 
Traditionally, the liver and liver lesion are delineated by radiologists on a slice-by-slice basis, which is time-consuming and prone to inter- and intra-rater variations. 
Therefore, automatic liver and liver tumor segmentation methods are highly demanded in the clinical practice.  

\begin{figure}[t]
	\centering
	\includegraphics[width=0.98\linewidth]{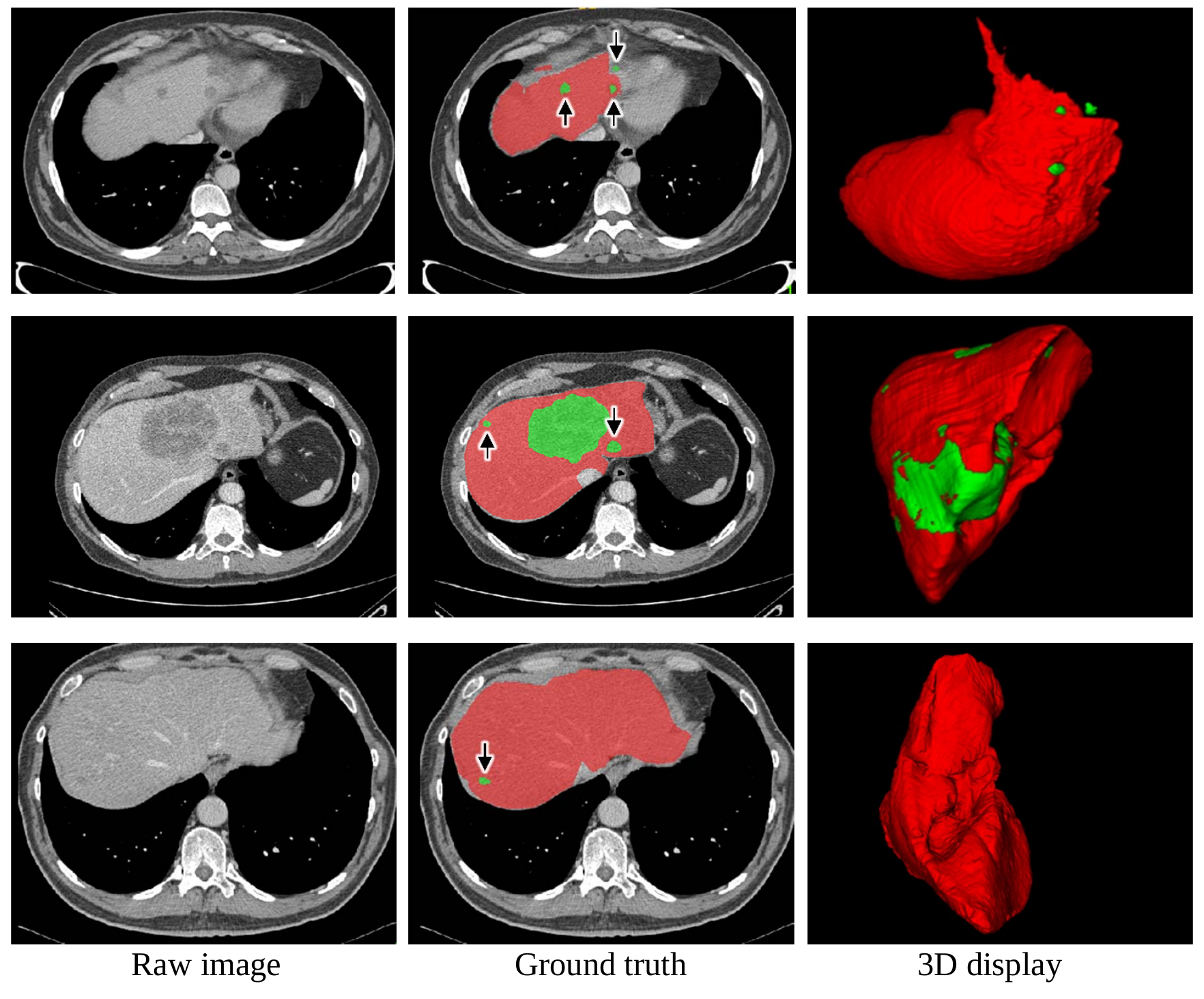}
	\caption{Examples of contrast-enhanced CT scans showing the large variations of shape, size, location of liver lesion. Each row shows a CT scan acquired from individual patient. The \emph{red} regions denote the liver while the \emph{green} ones denote the lesions (see the black arrows above).}
	\label{fig:challenges}\centering
\end{figure}

Automatic liver segmentation from the contrast-enhanced CT volumes is a very challenging task due to the low intensity contrast between the liver and other neighboring organs (see the first row in Figure~\ref{fig:challenges}). Moreover, radiologists usually enhance CT scans by an injection protocol for clearly observing tumors, which may increase the noise inside the images on the liver region~\cite{moghbel2017review}. Compared with liver segmentation, liver tumor segmentation is considered to be a more challenging task. First, the liver tumor has various size, shape, location and numbers within one patient, which hinders the automatic segmentation, as shown in Figure~\ref{fig:challenges}. Second, some lesions do not have clear boundaries, limiting the performance of solely edge based segmentation methods (see the lesions in the third row of Figure~\ref{fig:challenges}).
Third, many CT scans consist of anisotropic dimensions with high variations along the $z$-axis direction (the voxel spacing ranges from 0.45mm to 6.0mm), which further poses challenges for automatic segmentation methods. 

To tackle these difficulties, many segmentation methods have been proposed, including intensity thresholding, region growing, and deformable models. These methods,
however, rely on hand-crafted features, and have limited feature
representation capability.
Recently, fully convolutional neural networks (FCNs) have achieved great success on a broad array of recognition problems~\citep{prasoon2013deep, ronneberger2015u,stollenga2015parallel, roth2015deeporgan,wang2015detection, li2016multi, cciccek20163d, havaei2017brain,chen2017voxresnet,wang2017liver,li20183d}. 
Many researchers advance this stream using deep learning methods in the liver and tumor segmentation problem and the literature can be classified into two categories broadly.
(1) 2D FCNs, such as UNet architecture~\citep{christ2016automatic}, the multi-channel FCN~\citep{sun2017automatic}, and the FCN based on VGG-16~\citep{ben2016fully}.
(2) 3D FCNs, where 2D convolutions are replaced by 3D convolutions with volumetric data input~\citep{dou20163d,lu2017automatic}.

\revise{In the clinical diagnosis, the experienced radiologist usually observes and segments tumors according to many adjacent slices along the $z$-axis.}
However, 2D FCN based methods ignore the contexts on the $z$-axis, which would lead to limited segmentation accuracy. To be specific, single or three adjacent slices cropped from volumetric images are fed into 2D FCNs~\citep{sun2017automatic,ben2016fully} and the 3D segmentation volume is generated by simply stacking the 2D segmentation maps. Although adjacent slices are employed, it is still not enough to probe the spatial information along the third dimension, which may degrade the segmentation performance.
To solve this problem, some researchers proposed to use tri-planar schemes or RNN to probe the 3D contexts~\citep{prasoon2013deep, tseng2017joint, cai2017improving}.
For example, ~\citet{prasoon2013deep} applied three 2D FCNs on orthogonal planes (e.g., the $xy$, $yz$, and $xz$ planes) and voxel prediction results are generated by the average of these probabilities.
\revise{Compared to 2D FCNs, 3D FCNs suffer from high computational cost and GPU memory consumption.}
The high memory consumption limits the depth of the network as well as the filter's field-of-view, which are the two key factors for performance gains~\citep{simonyan2014very}.
\revise{The heavy computation of 3D convolutions also impedes the application in training  a large-scale dataset.}
Moreover, many researchers have demonstrated the effectiveness of knowledge transfer (the knowledge learnt from one source
domain efficiently transferred to another domain) for boosting the performance~\citep{chen2015standard,tajbakhsh2016convolutional}.
Unfortunately, only a dearth of 3D pre-trained model exists, which restricts the performance and also the adoption of 3D FCNs.

To address the above problems, we proposed a novel end-to-end system, called hybrid densely connected UNet (H-DenseUNet), where intra-slice features and 3D contexts are effectively probed and jointly optimized for accurate liver and lesion segmentation.
Our H-DenseUNet has the following two technical achievements:

\revise{\textbf{Deep and efficient network.} \
	First, to fully extract high-level intra-slice features, we design a very deep and efficient network based on the pre-defined design principles by 2D convolutions, called 2D DenseUNet, where the advantages of both densely connected path~\citep{huang2017densely} and UNet connections~\citep{ronneberger2015u} are fused together.
	Densely connected path is derived from densely connected network (DenseNet), where the improved information flow and parameters efficiency alleviate the difficulty for training the deep network.
	Different from DenseNet~\citep{huang2017densely}, we add the UNet connections, i.e., long-range skip connections, between the encoding part and the decoding part in our architecture; hence, the network can enable low-level spatial feature preservation for better intra-slice context exploration.}

\textbf{Hybrid feature exploration.} \
Second, to explore the volumetric feature representation, we design an end-to-end training system, called H-DenseUNet, where intra-slice and inter-slice features are effectively extracted and then jointly optimized through the hybrid feature fusion (HFF) layer.  
Specifically, 3D DenseUNet is integrated with the 2D DenseUNet by the way of auto-context~\citep{tu2008auto} mechanism, which is a general form of stacked generality~\citep{wolpert1992stacked}.
With the guidance of semantic probabilities from 2D DenseUNet, the optimization burden in the 3D DenseUNet can be well alleviated, which contributes to the training efficiency for 3D contexts extraction.
Moreover, with the end-to-end system, the hybrid feature, consisting of volumetric features and the high-level representative intra-slice features, can be automatically fused and jointly optimized together for better liver and tumor recognition.
In summary, this work has the following achievements:
\begin{itemize}
	\item We design a DenseUNet to effectively probe hierarchical intra-slice features for liver and tumor segmentation, where the densely connected path and UNet connections are carefully integrated \revise{ based on pre-defined design principles} to improve the liver tumor segmentation performance.
	
	\item We propose a H-DenseUNet framework to explore hybrid (intra-slice and inter-slice) features for liver and tumor segmentation. \revise{The hybrid feature learning architecture well sidesteps the problems that 2D networks neglect the volumetric contexts and 3D networks suffer from heavy computational cost, and can be served as a new paradigm for effectively exploiting 3D contexts. }
	
	\item 
	\revise{Our method ranked the 1st on lesion segmentation, achieved very competitive performance on liver segmentation in the  2017 LiTS Leaderboard, and also achieved the state-of-the-art results on the 3DIRCADb Dataset.}
\end{itemize}


\section{Related Work}
\subsection{Hand-crafted feature based methods} In the past decades, a lot of algorithms, including thresholding~\citep{soler2001fully,moltz2008segmentation}, region growing, deformable model based methods~\citep{wong2008semi,jimenez2011optimal} and machine learning based methods~\citep{huang2014random,vorontsov2014metastatic,le2016liver,kuo2017texture,conze2017scale} have been proposed to segment liver and liver tumor. 
Threshold-based methods classified foreground and background according to whether the intensity value is above a threshold. 
Variations of region growing algorithms were also popular in the liver and lesion segmentation task. For example, \citet{wong2008semi} segmented tumors by a 2D region growing method with knowledge-based constraints. 
Level set methods also attracted attentions from researchers with the advantages of numerical computations involving curves and surfaces~\cite{hoogi2017adaptive}. For example, \citet{jimenez2011optimal} proposed to classify tumors by a multi-resolution 3D level set method coupled with adaptive curvature technique. 
A large variety of machine learning based methods have also been proposed for liver tumor segmentation. For example, \citet{huang2014random} proposed to employ the random feature subspace ensemble-based extreme  learning machine (ELM) for liver lesion segmentation. \citet{vorontsov2014metastatic} proposed to segment tumors by support vector machine (SVM) classifier and then refined the results by the omnidirectional deformable surface model. 
Similarly, \citet{kuo2017texture} proposed to learn SVM classifier with texture feature vector for liver tumor segmentation. 
\citet{le2016liver} employed the fast marching algorithm to generate initial regions and then classified tumors by training a noniterative single hidden layer feedforward network (SLFN).
To speed up the segmentation algorithm, \citet{chaieb2017accelerated} adopted a bootstrap sampling approach for efficient liver tumor segmentation.
\subsection{Deep learning based methods}

Convolutional neural networks (CNNs) have achieved great success in many object recognition problems in computer vision community.
Many researchers followed this trend and proposed to utilize various CNNs for learning feature representations in the application of liver and lesion segmentation.
For example, \citet{ben2016fully} proposed to use a FCN for liver segmentation and liver-metastasis detection in CT examinations. \citet{christ2016automatic,christ2017automatic} proposed a cascaded FCN architecture and dense 3D conditional random fields (CRFs) to automatically segment liver and liver lesions. In the meanwhile, \citet{sun2017automatic} designed a multi-channel FCN to segment liver tumors from CT images, where the probability maps were generated by the feature fusion from different channels.



Recently, during the 2017 ISBI LiTS challenge, \citet{han2017automatic}, proposed a 2.5D 24-layer FCN model to segment liver tumors, where the residual block was employed as the repetitive building blocks and the UNet connection was designed across the encoding part and decoding part. 
2.5D refers to using 2D convolutional neural network with the input of adjacent slices from the volumetric images. 
Both \citet{vorontsov2017liver} and \citet{chlebus2017neural} achieved the second place in the ISBI challenge. \citet{vorontsov2017liver} also employed ResNet-like residual blocks and UNet connections with 21 convolutional layers, which is a bit shallower and has fewer parameters compared to that proposed by \citet{han2017automatic}. \citet{chlebus2017neural} designed a 28-layer UNet architecture in two individual models and subsequently filtered the false positives of tumor segmentation results by a random forest classifier. 
Instead of using 3D FCNs, all of the top results employed 2D FCNs with different network depths, showing the efficacy of 2D FCNs regarding the underlying volumetric segmentation problem.
However, all these networks are shallow and ignore the 3D contexts, which limit the high-level feature extraction capability and restrict the recognition performance. 

\begin{figure*}[!t]
	\centering
	\includegraphics[width=0.98\linewidth]{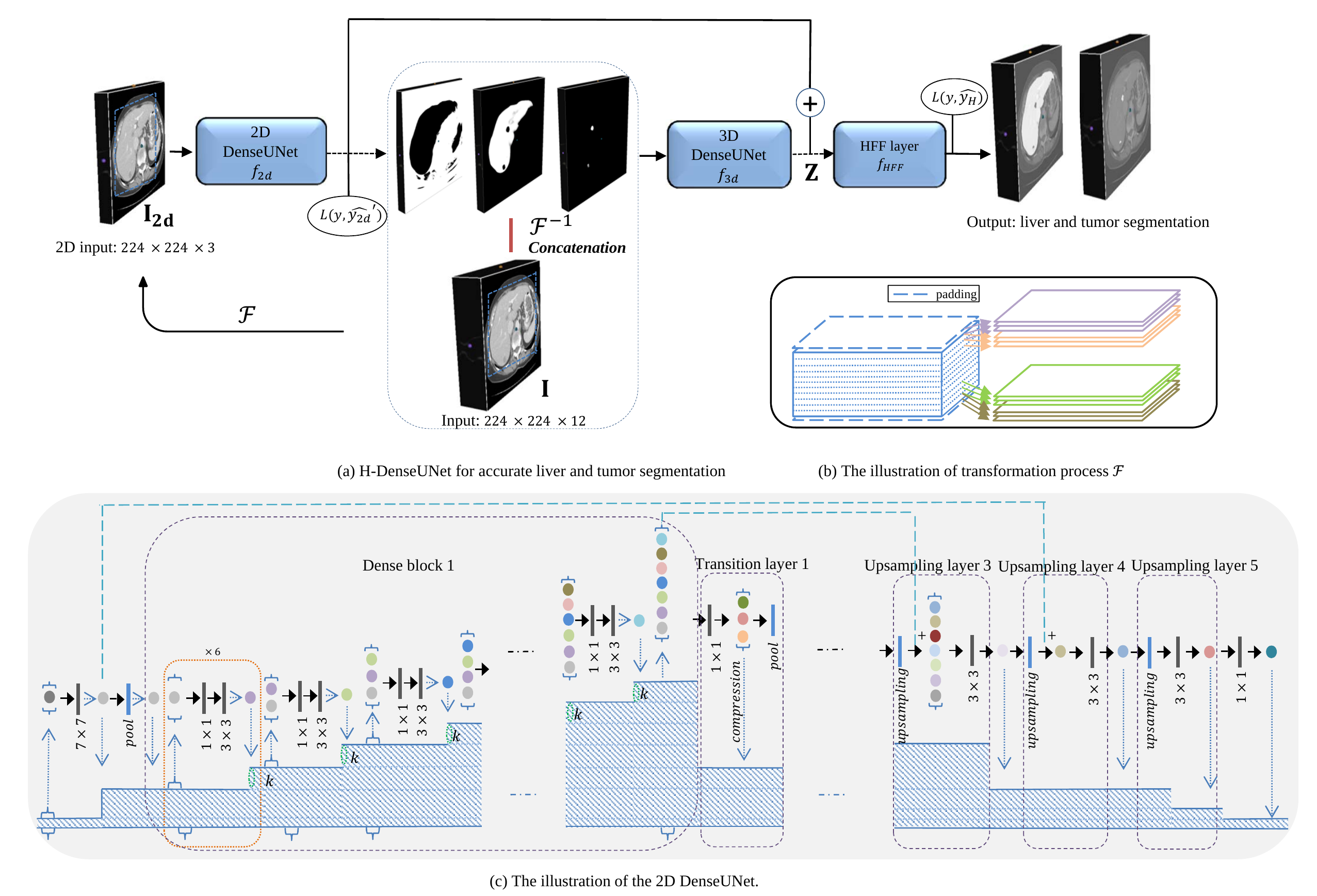}
	\caption{\revise{The illustration of the pipeline for liver and lesion segmentation. Each 3D input volume is sliced into adjacent slices through transformation process $\mathscr{F}$ and then fed into 2D DenseUNet; Concatenated with the prediction volumes from 2D network, the 3D input volumes are fed into the 3D network for learning inter-slice features; Then, the HFF layer fused and optimized the intra-slice and inter-slice features for accurate liver and tumor segmentation. 
			(a) The structure of H-DenseUNet, including the 2D DenseUNet and the 3D counterpart.
			(b) The transformation of the volumetric data to three adjacent slices. (c) The network structure of the 2D DenseUNet. The structure in the orange block is a micro-block and $k$ denotes the growth-rate. (Best viewed in color) }}
	\label{fig:pipeline}\centering
\end{figure*}

\section{Method}
Figure~\ref{fig:pipeline} shows the pipeline of our proposed method for liver and tumor segmentation.
We employed the cascaded learning strategy to reduce the overall computation time, which has also been adopted in many recognition tasks~\cite{zhou2017fixed,farag2017bottom,zhou2017deep,roth2017spatial}.
First, a simple ResNet architecture~\citep{han2017automatic} is trained to get a quick but coarse segmentation of liver.
With the region of interest (ROI), our proposed H-DenseUNet efficiently probes intra-slice and inter-slice features through a 2D DenseUNet $f_{2d}$ and a 3D counterpart $f_{3d}$, followed by jointly optimizing the hybrid features in the hybrid feature fusion (HFF) layer for accurate liver and lesion segmentation.

\subsection{Deep 2D DenseUNet for Intra-slice Feature Extraction}
The intra-slice feature extraction part follows the structure of DenseNet-161~\cite{huang2017densely}, which is composed of repetitive densely connected building blocks with different output dimensions.
In each densely connected building block, there are direct connections from any layer to all subsequent layers, as shown in Figure~\ref{fig:pipeline}(c).
Each layer produces $k$ feature maps and $k$ is called \emph{growth rate}. One advantage of the dense connectivity between layers is that it has fewer output dimensions than traditional networks, avoiding learning redundant features. Moreover, the densely connected path ensures the maximum information flow between layers, which improves the gradient flow, and thus alleviates the burden in searching for the optimal solution in a very deep neural network.

However, the original DenseNet-161~\cite{huang2017densely} is designed for the object classification task while our problem belongs to the segmentation topics. Moreover, a deep FCN network for segmentation tasks actually contains several max-pooling and upsampling operations, which may lead to the information loss of low-level (i.e., high resolution) features. Given above two considerations, we develop a 2D DenseUNet, which inherits both advantages of densely connected path and UNet-like connections~\citep{ronneberger2015u}. Specifically, the dense connection between layers is employed within each micro-block to ensure the maximum information flow while the UNet long range connection links the encoding part and the decoding part to preserve low-level information.

\begin{table*}[t]
	\centering \caption{Architectures of the proposed H-DenseUNet, consisting of the 2D DenseUNet and the 3D counterpart. The symbol --[ ] denotes the long range UNet summation connections with the last layer of the dense block. The second and forth column indicate the output size of the current stage in two architectures, respectively. Note that "$1\times1, 192$ conv" corresponds to the sequence BN-ReLU-Conv layer with convolutional kernel size of $1\times1$ and 192 features. "[ ]$ \times d$" represents the dense block is repeated for $d$ times.}
	\label{tab:densenet} %
	\resizebox{\textwidth}{!}{
		\begin{tabular}{c|c|c|c|c}
			\hline
			&  Feature size &  2D DenseUNet-167 (\emph{k}=48) & Feature size  & 3D DenseUNet-65 (\emph{k}=32) \tabularnewline
			\hline
			input & 224 $\times$ 224 & - & 224 $\times$ 224 $\times$ 12 & - \tabularnewline
			\hline
			convolution 1 & 112 $\times$ 112 & \revise{ 7 $\times$ 7, 96, stride 2} & 112 $\times$ 112 $\times$ 6 & \revise{7 $\times$ 7 $\times$ 7, 96, stride 2}
			\tabularnewline
			\hline
			pooling & 56 $\times$ 56 & \revise{3 $\times$ 3 max pool, stride 2}  & 56 $\times$ 56 $\times$ 3 & \revise{ 3 $\times$ 3 $\times$ 3 max pool, stride 2} \tabularnewline
			\hline
			dense block 1 & 56 $\times$ 56 & $ \begin{bmatrix}
			1 \times 1 , 192 &\text{conv}  \\
			3 \times 3 , 48 &\text{conv} &
			\end{bmatrix} \times 6$ &  56 $\times$ 56 $\times$ 3 & $ \begin{bmatrix}
			1 \times 1 \times 1, 128 &\text{conv}  \\
			3 \times 3 \times 3, 32 &\text{conv} &
			\end{bmatrix} \times 3$\tabularnewline
			\hline
			\multirow{2}*{transition layer 1} & 56 $\times$ 56 & 1 $\times$ 1 conv & 56 $\times$ 56 $\times$ 3 & 1 $\times$ 1 $\times$ 1 conv \tabularnewline
			\cline{2-5}
			& 28 $\times$ 28& 2 $\times$ 2 average pool & 28 $\times$ 28 $\times$ 3 & 2 $\times$ 2 $\times$ 1 average pool  \tabularnewline
			\hline
			dense block 2 & 28 $\times$ 28 & $ \begin{bmatrix}
			1 \times 1, 192 & \text{conv}  \\
			3 \times 3, 48 & \text{conv} &
			\end{bmatrix} \times 12$  & 28 $\times$ 28 $\times$ 3 & $ \begin{bmatrix}
			1 \times 1 \times 1, 128 & \text{conv}  \\
			3 \times 3 \times 3, 32 & \text{conv} &
			\end{bmatrix} \times 4$ \tabularnewline
			\hline
			\multirow{2}*{transition layer 2} & 28 $\times$ 28 & 1 $\times$ 1 conv & 28 $\times$ 28 $\times$ 3 & 1 $\times$ 1 $\times$ 1 conv  \tabularnewline
			\cline{2-5}
			& 14 $\times$ 14& 2 $\times$ 2 average pool, & 14 $\times$ 14 $\times$ 3 &   2 $\times$ 2 $\times$ 1 average pool \tabularnewline
			\hline
			dense block 3 & 14 $\times$ 14 & $ \begin{bmatrix}
			1 \times 1, 192 & \text{conv}  \\
			3 \times 3, 48 & \text{conv} &
			\end{bmatrix} \times 36$  & 14 $\times$ 14 $\times$ 3 & $ \begin{bmatrix}
			1 \times 1 \times 1, 128 & \text{conv}  \\
			3 \times 3 \times 3, 32 & \text{conv} &
			\end{bmatrix} \times 12$   \tabularnewline
			\hline
			\multirow{2}*{transition layer 3} & 14 $\times$ 14 & 1 $\times$ 1 conv & 14 $\times$ 14 $\times$ 3  & 1 $\times$ 1 $\times$ 1 conv \tabularnewline
			\cline{2-5}
			& 7 $\times$ 7 & 2 $\times$ 2 average pool & 7 $\times$ 7 $\times$ 3  & 2 $\times$ 2 $\times$ 1  average pool \tabularnewline
			\hline
			dense block 4 & 7 $\times$ 7 & $ \begin{bmatrix}
			1 \times 1, 192 & \text{conv}  \\
			3 \times 3, 48 & \text{conv} &
			\end{bmatrix} \times 24$  & 7 $\times$ 7 $\times$ 3 & $ \begin{bmatrix}
			1 \times 1 \times 1, 128 & \text{conv}  \\
			3 \times 3 \times 3, 32 & \text{conv} &
			\end{bmatrix} \times 8$ \tabularnewline
			\hline
			upsampling layer 1 & 14 $\times$ 14 & 2 $\times$ 2 upsampling -- [dense block 3], 768, conv  & 14 $\times$ 14 $\times$ 3  &  2 $\times$ 2 $\times$ 1 upsampling -- [dense block 3], 504, conv \tabularnewline
			\hline
			upsampling layer 2 & 28 $\times$ 28 & 2 $\times$ 2 upsampling -- [dense block 2], 384, conv &  28 $\times$ 28 $\times$ 3  &  2 $\times$ 2  $\times$ 1 upsampling -- [dense block 2], 224, conv  \tabularnewline
			\hline
			upsampling layer 3 & 56 $\times$ 56 & 2 $\times$ 2 upsampling -- [dense block 1], 96, conv & 56 $\times$ 56 $\times$ 3  &  2 $\times$ 2 $\times$ 1 upsampling -- [dense block 1], 192, conv  \tabularnewline
			\hline
			upsampling layer 4 & 112 $\times$ 112 & 2 $\times$ 2 upsampling -- [convolution 1], 96, conv & 112 $\times$ 112 $\times$ 6  & 2 $\times$ 2 $\times$ 2 upsampling -- [convolution 1], 96, conv \tabularnewline
			\hline
			upsampling layer 5 & 224 $\times$ 224 & 2 $\times$ 2 upsampling, 64, conv & 224 $\times$ 224 $\times$ 12  &   2 $\times$ 2 $\times$ 2 upsampling, 64, conv \tabularnewline
			\hline
			convolution  2 & 224 $\times$ 224 & 1 $\times$ 1, 3  & 224 $\times$ 224 $\times$ 12 &  1 $\times$ 1 $\times$ 1, 3 \tabularnewline
			\hline
		\end{tabular}
	}
\end{table*}
Let $\mathbf{I} \in R^{ n\times224\times224\times12\times1}$ denote the input training samples (for $224 \times 224 \times 12$ input volumes) with ground-truth labels $\mathbf{Y} \in R^{ n\times224\times224\times12\times1}$, where $n$ denotes the batch size of the input training samples and the last dimension denotes the channel. $\mathbf{Y}_{i,j,k}=c$ since each pixel ($i,j,k$) is tagged with class $c$ (background, liver and tumor). Let function $\mathscr{F}$ denote the transformation from the volumetric data to three adjacent slices. Specifically, every three adjacent slices along $z$-axis are stacked together and the number of groups can be transformed to the batch dimension.
For example, $\mathbf{I_{2d}} = \mathscr{F}(\mathbf{I})$, where $\mathbf{I_{2d}} \in R^{12n \times 224 \times 224 \times 3}$ denotes the input samples of 2D DenseUNet.
The detailed transformation process is illustrated in Figure~\ref{fig:pipeline}(d).
Because of the transformation, the 2D and 3D DenseUNet can be jointly trained, which will be described in detail in section B.
For convenience,  $\mathscr{F}^{-1}$ denotes the inverse transformation from three adjacent slices to the volumetric data.
The 2D DenseUNet conducts liver and tumor segmentation,
\begin{equation}
\begin{split}
\vspace{-1mm}
\label{eq:1}
\mathbf{X_{2d}} = f_{2d}(\mathbf{I_{2d}}; \theta_{2d}), \mathbf{X_{2d}}\in R^{12n \times 224\times 224 \times 64},\\
\hat{y_{2d}} = f_{2dcls}(\mathbf{X_{2d}}; \theta_{2dcls}), \hat{y_{2d}}\in R^{12n \times 224 \times 224 \times 3}
\end{split}
\end{equation}
where $\mathbf{X_{2d}}$ is the feature map from
layer "upsampling layer 5" (see Table~\ref{tab:densenet}) and $\hat{y_{2d}}$ is the corresponding pixel-wise probabilities for input $\mathbf{I_{2d}}$.

The illustration and detailed structure of 2D DenseUNet are shown in Figure~\ref{fig:pipeline}(c) and Table~\ref{tab:densenet}, respectively.
The depth of 2D DenseUNet is extended to 167 layers, referred as 2D DenseUNet-167, which consists of 167 convolution layers, pooling layers, dense blocks, transition layers and upsampling layers.
The dense block denotes the cascade of several micro-blocks, in which all layers are directly connected, see Figure~\ref{fig:pipeline}(c).
To change the size of feature-maps, the transition layer is employed, which consists of a batch normalization layer and a $1 \times 1$ convolution layer followed by an average pooling layer. A compression factor is included in the transition layer to compress the number of feature-maps, preventing the expanding of feature-maps (set as 0.5 in our experiments).
The upsampling layer is implemented by the bilinear interpolation, followed by the summation with low-level features (i.e., UNet connections) and a $3 \times 3$ convolutional layer.
Before each convolution layer, the batch normalization and the Rectified Linear Unit (ReLU) are employed in the architecture.

\subsection{H-DenseUNet for Hybrid Feature Exploration}
2D DenseUNet with deep convolutions can produce high-level representative in-plane features but neglect the spatial information along the $z$ dimension while 3D DenseUNet has large GPU computational cost and limited kernel's field-of-view as well as the network depth.
To address these issues, we propose H-DenseUNet to jointly fuse and optimize the learned intra-slice and inter-slice features for better liver tumor segmentation.


To fuse hybrid features from the 2D and 3D network, the feature volume size should be aligned. Therefore, the feature maps and score maps from 2D DenseUNet are transformed to the volumetric shape as follows:
\begin{equation}
\begin{split}
\vspace{-1mm}
\label{eq:3}
\mathbf{X_{2d}}' =\mathscr{F}^{-1}(\mathbf{X_{2d}}), \mathbf{X_{2d}}' \in R^{n \times 224 \times 224 \times 12 \times 64},\\
\hat{y_{2d}}' = \mathscr{F}^{-1}(\hat{y_{2d}}), \hat{y_{2d}}' \in R^{n \times 224 \times 224 \times 12 \times 3},\\
\end{split}
\end{equation}
Then the 3D DenseUNet distill the visual features with 3D contexts by concatenating the original volumes $\mathbf{I} $ with the contextual information $\hat{y_{2d}}'$ from the 2D network.
Specifically, the detectors in the 3D counterpart trained based not only on the features probed from the original images, but also on the probabilities of a large number of context pixels from 2D DenseUNet. 
With the guidance from the supporting contexts pixels, the burden in searching for the optimal solution in the 3D counterpart has also been well alleviated, which significantly improves the learning efficiency of the 3D network.
The learning process of 3D DenseUNet can be described as:
\begin{equation}
\begin{split}
\vspace{-1mm}
\label{eq:2}
\mathbf{X_{3d}} =  f_{3d}(\mathbf{I},\hat{y_{2d}}';\theta_{3d}), \\
\mathbf{Z} = \mathbf{X_{3d}} + \mathbf{X_{2d}}', \\
\end{split}
\end{equation}
where $\mathbf{X_{3d}}$ denotes the feature volume from layer "\emph{upsampling layer 5}" in 3D DenseUNet-65. $\mathbf{Z}$ denotes the hybrid feature, which refers to the sum of intra-slice and inter-slice features from 2D and 3D network, respectively. Then the hybrid feature is jointly learned and optimized in the HFF layer,
\begin{equation}
\begin{split}
\vspace{-1mm}
\label{eq:5}
\mathbf{H} = f_{HFF}(\mathbf{Z};\theta_{HFF}),\\
\hat{y_{H}} = f_{HFFcls}(\mathbf{H}; \theta_{HFFcls})
\end{split}
\end{equation}
where H denotes the optimized hybrid features and $\hat{y_{H}}$ refers to the pixel-wise predicted probabilities generated from the HFF layer $f_{HFFcls}(\cdot)$.
In our experiments, the 3D counterpart of H-DenseUNet cost only 9 hours to converge, which is significantly faster than training the 3D counterpart with original data solely (63 hours).

The detailed structure of the 3D counterpart is shown in the Table~\ref{tab:densenet}, called 3D DenseUNet-65, which consists of 65 convolutional layers and the growth rate is 32.
Compared with 2D DenseUNet counterpart, the number of micro-blocks in each dense block is decreased due to the high memory consumption of 3D convolutions and the limited GPU memory.
The rest of the network setting is the same with the 2D counterpart.

\subsection{Loss Function, Training and Inference Schemes}
In this section, we present more details regarding the loss function, training and the inference schemes.

\subsubsection{Loss Function}
To train the networks, we employed weighted cross-entropy function as the loss function, which is described as:
\begin{equation}
\vspace{-1mm}
\label{eq: weighted loss}
L(y,\hat{y}) = -\frac{1}{N}\sum_{i=1}^{N}\sum_{c=1}^{3}w_i^c y_i^c \log \hat{y_i}^c
\end{equation}
where $\hat{y_i}^c$ denotes the probability of voxel $i$ belongs to class $c$ (background, liver or lesion), $w_i^c$ denotes the weight and $y_i^c$ indicates the ground truth label for voxel $i$. 

\subsubsection{Training Scheme}
We first train the ResNet in the same way with~\citet{han2017automatic} to get the coarse liver segmentation results.
The parameters of the encoder part in 2D DenseUNet $f_{2d}$ are initialized with DenseNet's weights (object classification-trained)~\cite{huang2017densely} while the decoder part are trained with the random initialization.
Since the weights are initialized with a random distribution in the decoder part, we first warm up the network without UNet connections.
After several iterations, the UNet connections are added to jointly fine tune the model.

To effectively train the H-DenseUNet, we first optimize $f_{2d}(\cdot)$ and $f_{2dcls}(\cdot)$ with cross entropy loss $L(y,\hat{{y_{2d}}}')$ on our dataset. 
Secondly, we fix parameters in $f_{2d}(\cdot)$ and $f_{2dcls}(\cdot)$, and focus on training $f_{3d}(\cdot), f_{HFF}(\cdot)$ and $ f_{HFFcls}(\cdot)$ with cross entropy loss $ L(y,\hat{{y_H}})$, where parameters are all randomly initialized. Finally, The whole network is jointly fine-tuned with following combined loss:
\begin{equation}
\begin{split}
\vspace{-1mm}
\label{eq:4}
L_{total} = \lambda L(y,\hat{{y_{2d}}}') + L(y,\hat{{y_H}})
\end{split}
\end{equation}
where $\lambda$ is the balanced weight and set as 0.5 in our experiments empirically. 

\subsubsection{Inference Scheme}
In the test stage, we first get the coarse liver segmentation result. Then H-DenseUNet can generate accurate liver and tumor predicted probabilities within the ROI.
The thresholding is applied to get the liver tumor segmentation result.
To avoid the holes in the liver, a largest connected component labeling is performed to refine the liver result.
After that, the final lesion segmentation result is obtained by removing lesions outside the final liver region.


\section{Experiments and Results}
\subsection{Dataset and Pre-processing}
We tested our method on the competitive dataset of MICCAI 2017 LiTS Challenge and 3DIRCADb Dataset. The LiTS dataset contains 131 and 70 contrast-enhanced 3D abdominal CT scans for training and testing, respectively.
The dataset was acquired by different scanners and protocols from six different clinical sites, with a largely varying in-plane resolution from 0.55 mm to 1.0 mm and slice spacing from 0.45 mm to 6.0 mm.
The 3DIRCADb dataset contains 20 venous phase enhanced CT scans, where 15 volumes have hepatic tumors in the liver.

For image preprocessing, we truncated the image intensity values of all scans to the range of [-200,250] HU to remove the irrelevant details.
For coarse liver segmentation in the first stage, we trained a simple network from resampled images with the same resolution $ 0.69 \times 0.69 \times 1.0~\rm{mm}^3$. In the test stage, we also employ the resampled images for coarse  liver segmentation. For lesion segmentation in the second stage, the network is trained on the images with the original resolution. This is because in some training cases liver lesions are notably small, thus we use images with the original resolution to avoid possible artifacts from image resampling. In this test stage, we also employ the images with original resolution for accurate liver and lesion segmentation.

\subsection{Evaluation Metrics}
According to the evaluation of 2017 LiTS challenge, we employed Dice per case score and Dice global score to evaluate the liver and tumor segmentation performance respectively.
Dice per case score refers to an average Dice score per volume while Dice global score is the Dice score evaluated by combining all datasets into one.
Root mean square error (RMSE) is also adopted to measure the tumor burden.

\revise{In the 3DIRCADb dataset, 
	five metrics are used 
	to measure the accuracy of segmentation results, including the volumetric overlap error (VOE), relative volume difference (RVD), average symmetric surface distance (ASD), root mean square symmetric surface distance (RMSD) and DICE. For the first four evaluation metrics, the smaller the value is, the better the segmentation result. The value of DICE refers to the same measurement as Dice per case in the LiTS dataset.}

\subsection{Implementation Details}
In this section, we present more details regarding the implementation environment and data augmentation strategies.
The model was implemented using \emph{Keras} package~\citep{chollet2015keras}.
The initial learning rate was 0.01 and decayed according to the equation $lr = lr * (1- iterations/total\_iterations)^ {0.9}$. We used stochastic gradient descent with momentum.

For data augmentation, we adopted random mirror and scaling between 0.8 and 1.2 for all training data to alleviate the overfitting problem.
The training of 2D DenseUNet model took about 21 hours using two NVIDIA Titan Xp GPUs with 12 GB memory while the end-to-end system fine-tuning cost approximately 9 hours.
In other words, the total training time for H-DenseUNet took about 30 hours.
In the test phase, the total processing time of one subject depends on the number of slices, ranging from 30 seconds to 200 seconds.

\begin{table*}[!t]
	\centering
	\caption{Segmentation results by ablation study of our methods on the test dataset (Dice: \%).}
	\label{tab:ablation study} %
	{	
		\begin{tabular}{c|c|c|c|c}
			\hline
			\multirow{2}{*}{Model}	& \multicolumn{2}{c|}{ Lesion } & \multicolumn{2}{|c}{ Liver }  \tabularnewline
			\cline{2-5}
			& Dice per case & Dice global   &  Dice per case  & Dice global  \tabularnewline
			\hline
			3D DenseUNet without pre-trained model & 59.4 &  78.8  & 93.6  & 92.9     \tabularnewline
			\hline

			UNet~\citep{chlebus2017neural} & 65.0 & -   &  - &  -  \tabularnewline
			\hline
			ResNet~\citep{han2017automatic} & 67.0 & -  &  - &  - \tabularnewline
			\hline
			2D DenseUNet without pre-trained model  & 67.7  &  80.1  &  94.7 &  94.7 \tabularnewline
			\hline
			2D DenseNet with pre-trained model  & 68.3 &  81.8 &  95.3  &  95.9  \tabularnewline
			\hline
			2D DenseUNet with pre-trained model & 70.2 & 82.1   & 95.8 & 96.3  \tabularnewline
			\hline
			H-DenseUNet  & \textbf{72.2} & \textbf{82.4} & \textbf{96.1}  &  \textbf{96.5}  \tabularnewline
			\hline	 
		\end{tabular}
	}
\end{table*}
\begin{table*}[htpb]
	\centering \caption{Leaderboard of 2017 Liver Tumor Segmentation (LiTS) Challenge (Dice: \%, until 1st Nov. 2017)}
	\label{tab:result} %
	{
		\centering
		\begin{tabular}{c|c|c|c|c|c}
			\hline
			\multirow{2}{*}{Team} & \multicolumn{2}{c|}{ Lesion } & \multicolumn{2}{|c}{ Liver }  & \multicolumn{1}{|c}{ Tumor Burden }\tabularnewline
			\cline{2-6}
			& Dice per case	& Dice global    & Dice per case & Dice global & RMSE \tabularnewline
			\hline
			\textbf{our} &  \textbf{72.2} & \textbf{82.4} & \textbf{96.1} & \textbf{96.5} &   \textbf{0.015}  \tabularnewline
			\hline
			IeHealth & 70.2 & 79.4  & \textbf{96.1} & 96.4   & 0.017 \tabularnewline
			\hline
			
			hans.meine & 67.6 & 79.6 & 96.0 & \textbf{96.5}  &  0.020 \tabularnewline
			\hline
			superAI & 67.4 & 81.4  & 0.0 & 0.0  & 1251.447 \tabularnewline
			\hline
			Elehanx \citep{han2017automatic}&  67.0 & -   & -  & - &  - \tabularnewline
			\hline
			medical & 66.1 & 78.3  & 95.1 & 95.1 &  0.023  \tabularnewline
			\hline
			deepX~\citep{yuan2017hierarchical} & 65.7 & 82.0 & \textbf{96.3} & \textbf{96.7}  & 0.017 \tabularnewline
			\hline
			Njust768 & 65.5  & 76.8  & 4.10 & 13.5   & 0.920 \tabularnewline
			\hline
			Medical \citep{vorontsov2017liver}&  65.0  & -  & -  & - & - \tabularnewline
			\hline
			Gchlebus \citep{chlebus2017neural} &  65.0 & -   & -  & -  & - \tabularnewline
			\hline
			predible & 64.0 & 77.0 & 95.0 &  95.0   & 0.020 \tabularnewline
			\hline	
			Lei \citep{bi2017automatic} &  64.0 &  - & -  & - & - \tabularnewline
			\hline
			ed10b047 & 63.0 & 77.0   & 94.0 & 94.0 &  0.020  \tabularnewline
			\hline	
			chunliang & 62.5 & 78.8 & 95.8 & 96.2  &   0.016 \tabularnewline
			\hline
			yaya  &  62.4 & 79.2  & 95.9 & 96.3  & 0.016 \tabularnewline
			\hline
		\end{tabular}
		
		{	
			\begin{tablenotes}
				\centering
				\small
				\item Note: - denotes that the team participated in ISBI competition and the measurement was not evaluated.
			\end{tablenotes}
		}
	}
\end{table*}
\subsection{Ablation Analysis of H-DenseUNet on LiTS dataset}
\begin{figure}[!t]
	\centering
	\includegraphics[width=1.0\linewidth]{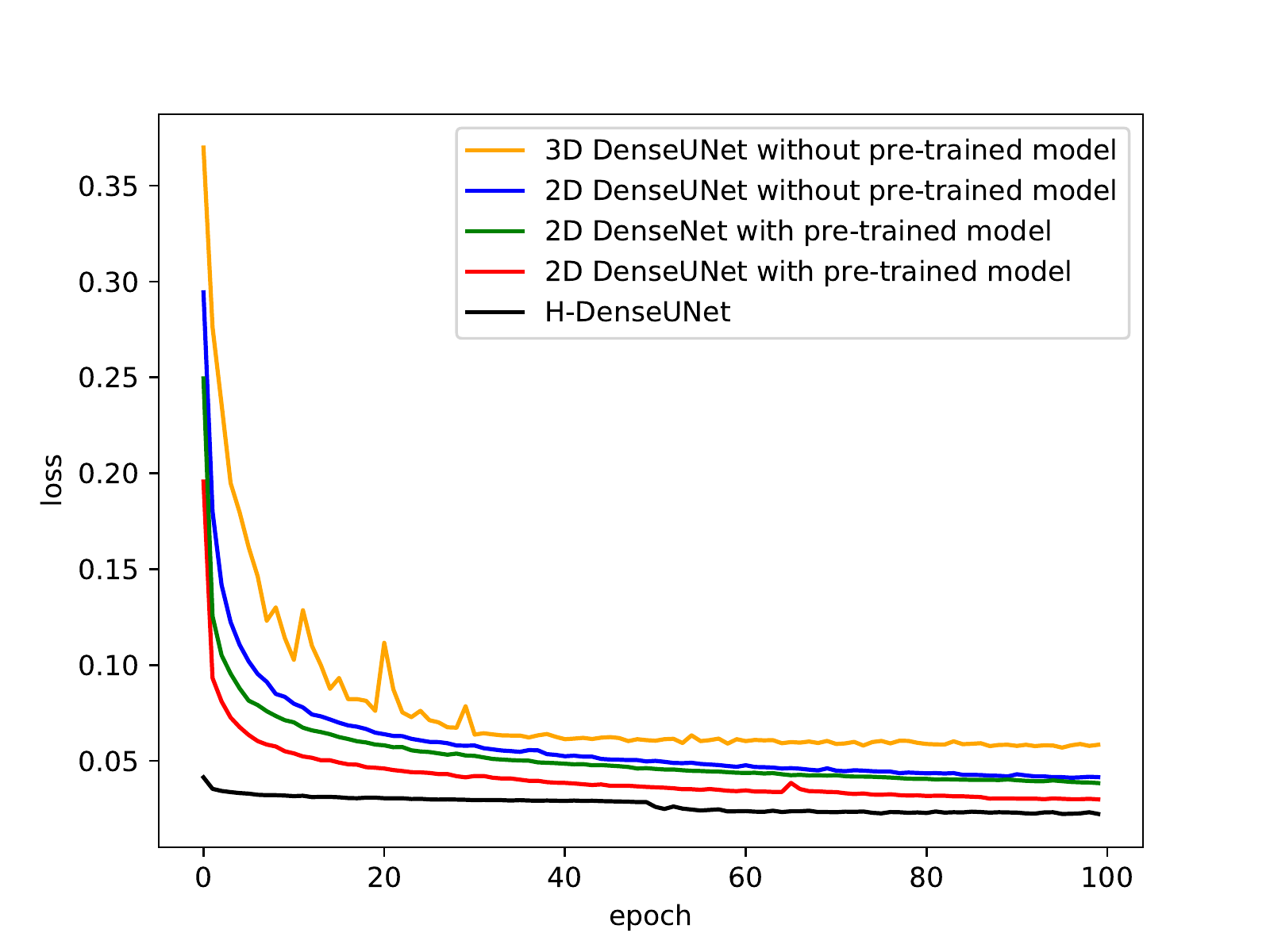}
	\caption{Training losses of 2D DenseUNet with and without pre-trained model, 2D DenseNet with pre-trained model, 3D DenseUNet without pre-trained model as well as H-DenseUNet (Best viewed in color).}
	\label{fig:loss}\centering
\end{figure}

In this section, we conduct comprehensive experiments to analyze the effectiveness of our proposed H-DenseUNet.
Figure~\ref{fig:loss} shows the training losses of 2D DenseUNet with and without pre-trained model, 2D DenseNet with pre-trained model, 3D DenseUNet without pre-trained model as well as H-DenseUNet. Note that 3D DenseUNet costs around 60 hours, nearly 3 times than 2D networks. H-DenseUNet costs nearly 30 hours, where 21 hours are spent for 2D DenseUNet training and 9 hours are used to fine-tune the whole architecture in the end-to-end manner. It is worth mentioning that all of the models are run with NVIDIA Titan Xp GPUs with full memory.
\subsubsection{Effectiveness of the Pre-trained Model}
One advantage in the proposed method is that we can train the network by transfer learning with the pre-trained model, which is crucial in finding an optimal solution for the network. Here, we analyze the learning behaviors of 2D DenseUNet with and without the pre-trained model. Both two experiments were conducted under the same experimental settings. From Figure~\ref{fig:loss}, it is clearly observed that with the pre-trained model, 2D DenseUNet can converge faster and achieve lower loss value, which shows the importance of utilizing the pre-trained model with transfer learning. The test results in Table~\ref{tab:ablation study} demonstrated that the pre-trained model can help the network achieve better performance consistently. Our proposed H-DenseUNet inherits this advantage, which plays an important role in achieving the promising results.
\subsubsection{Comparison of 2D and 3D DenseUNet}
We compare the inherent performance of 2D DenseUNet and 3D DenseUNet to validate that using 3D network solely maybe defective.
The number of parameters is one of key elements in measuring the model representation capability, thus both 2D DenseUNet-167 and 3D DenseUNet-65 are designed with the same level of model complexity (around 40M parameters).

We compare the learning behaviors of two experiments without using the pre-trained model. From Figure~\ref{fig:loss}, it shows that the 2D DenseUNet achieves better performance than the 3D DenseUNet, which highlights the effectiveness and efficiency of 2D convolutions with the deep architecture. This is because the 3D kernel consumes large GPU memory so that the network depth and width are limited, leading to weak representation capability.
In addition, 3D DenseUNet took much more training time (approximately 60 hours) to converge compared to 2D DenseUNet (around 20 hours).

Except for the heavy computational cost of the 3D network, another defective is that only a dearth of pre-trained model exists for the 3D network.
From Table~\ref{tab:ablation study}, compared with the results generated by 3D DenseUNet, 2D DenseUNet with pre-trained model achieved 8.9 and 3.0 (Dice: \%) improvements on the lesion segmentation results by the measurement of Dice per case and Dice global score, respectively. 
\subsubsection{Effectiveness of UNet Connections}

We analyze the effectiveness of UNet connections in our proposed framework. Both 2D DenseNet and DenseUNet are trained with the same pre-trained model and training strategies. The difference is that DenseUNet contains long range connections between the encoding part and the decoding part to preserve high-resolution features. As the results shown in Figure~\ref{fig:loss}, it is obvious that DenseUNet achieves lower loss value than DenseNet, demonstrating the UNet connections actually help the network converge to a better solution. The experimental results in Table~\ref{tab:ablation study} consistently demonstrated that the lesion segmentation performance can be boosted by a large margin with UNet connections embedded in the network.
\begin{figure}[!t]
	\centering
	\includegraphics[width=1\linewidth]{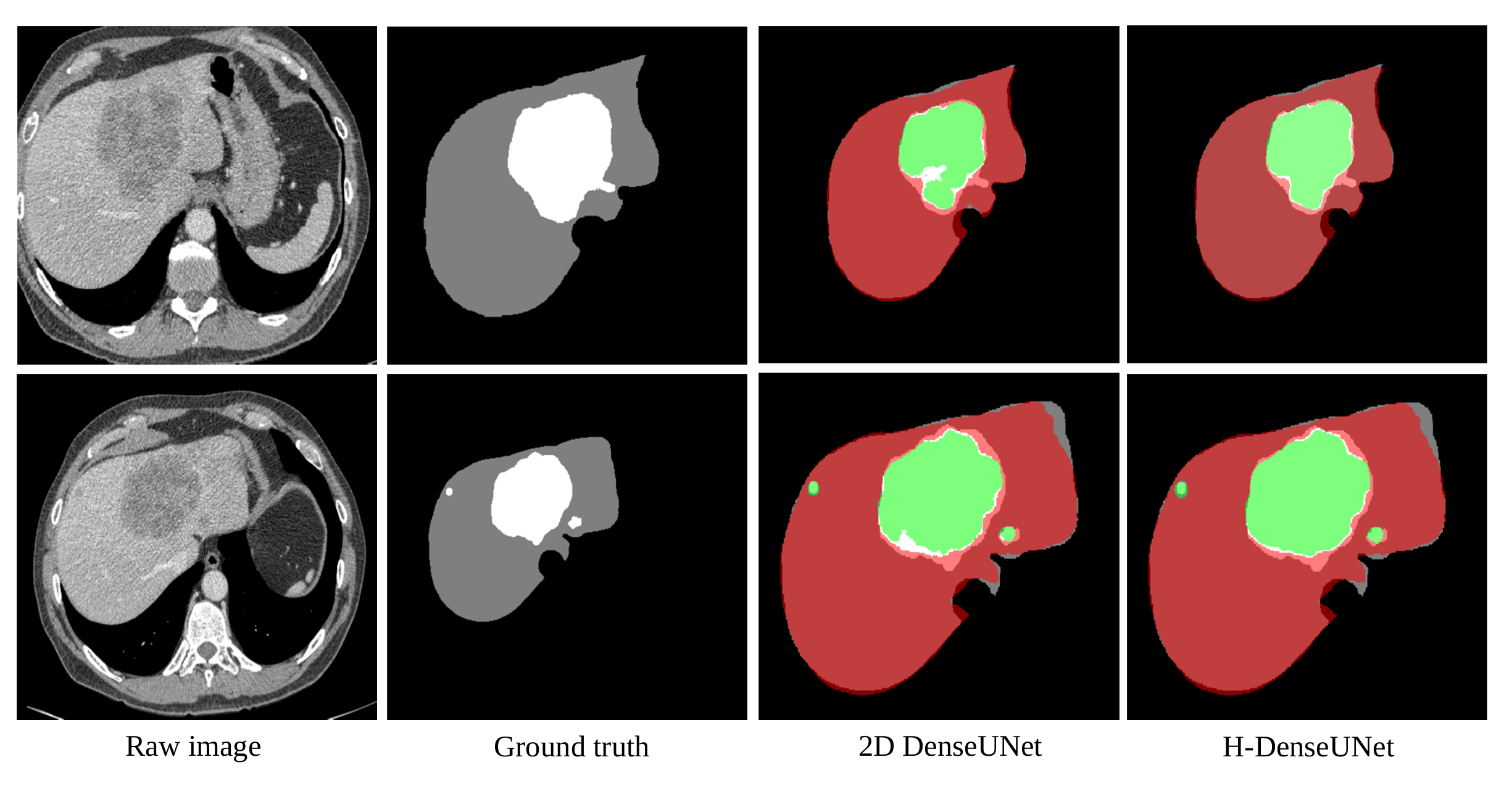}
	\caption{Examples of segmentation results by 2D DenseUNet and H-DenseUNet on the validation dataset. The \emph{red} regions denote the segmented liver while the \emph{green} ones denote the segmented lesions. The \emph{gray} regions denote the true liver while the \emph{white} ones denote the true lesions.}
	\label{fig:exam}\centering
\end{figure}
\subsubsection{Effectiveness of Hybrid Feature Fusion}
\begin{figure}[!t]
	\centering
	\includegraphics[width=0.98\linewidth]{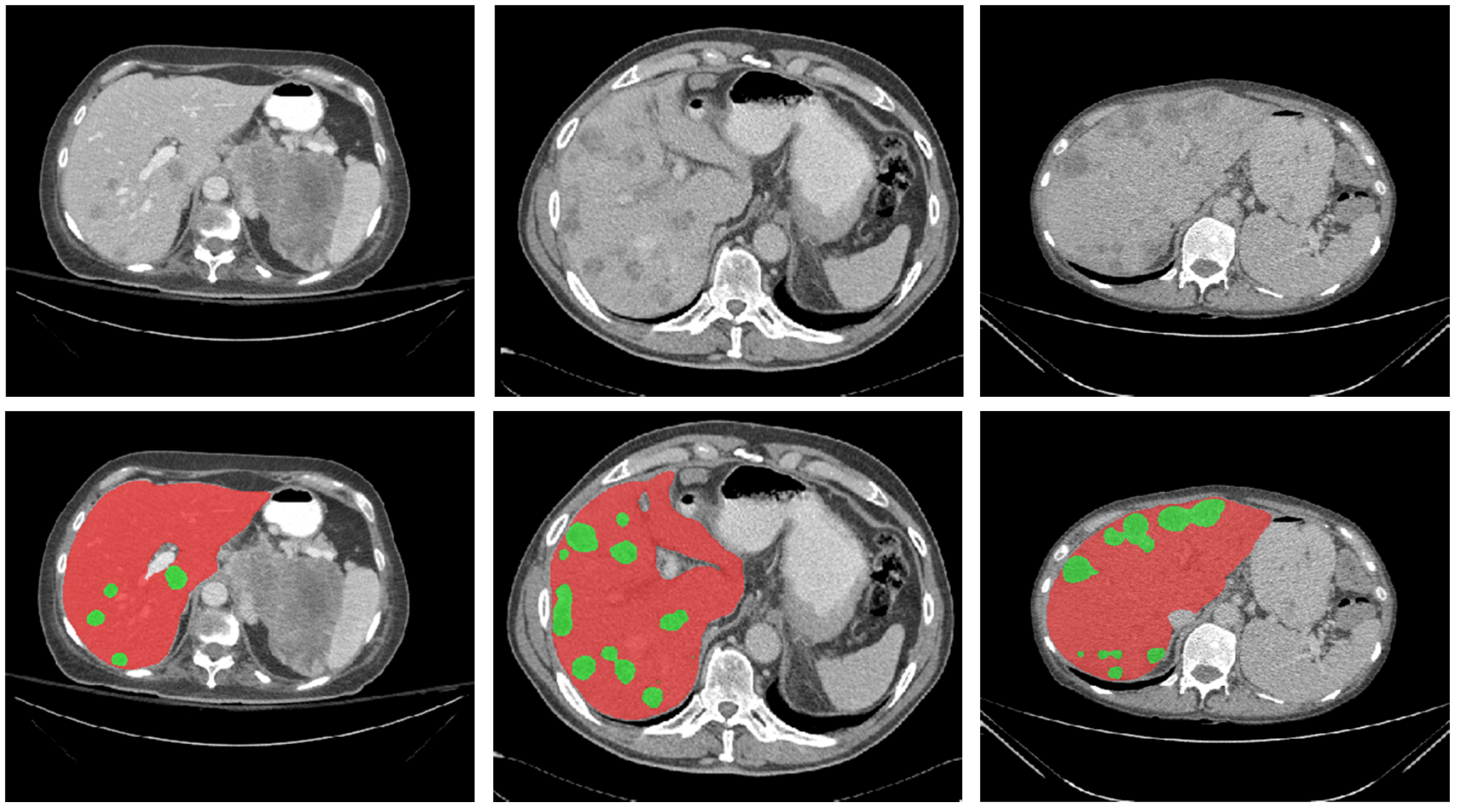}
	\caption{Examples of liver and tumor segmentation results of H-DenseUNet from the test dataset. The \emph{red} regions denote the liver and the \emph{green} ones denote the tumors.}
	\label{fig:result}\centering
\end{figure}

To validate the effectiveness of the hybrid architecture, we compare the learning behaviors of H-DenseUNet and 2D DenseUNet.
It is observed that the loss curve for H-DenseUNet begins around 0.04. This is because we fine tune the H-DenseUNet on the 2D DenseUNet basis, which serves as a good initialization.
Then the loss value decreases to nearly 0.02, which is attributed to the hybrid feature fusion learning.
Figure~\ref{fig:loss} shows that H-DenseUNet can converge to the smaller loss value than the 2D DenseUNet, which indicates that the hybrid architecture can contribute to the performance gains.
Compared with 2D DenseUNet, our proposed H-DenseUNet advances the segmentation results on both two measurements for liver and tumor segmentation consistently, as shown in Table~\ref{tab:ablation study}.
The performance gains indicate that contextual information along the $z$ dimension, indeed, contributes to the recognition of lesion and liver, especially for lesions that have much more blurred boundary and considered to be difficult to recognize.
Figure~\ref{fig:exam} shows some segmentation results achieved by 2D DenseUNet and H-DenseUNet on the validation dataset. It is observed that H-DenseUNet can achieve much better results than 2D DenseUNet.
Moreover, we trained H-DenseUNet in an end-to-end manner, where the 3D contexts can also help extract more representative in-plane features.
The end-to-end system jointly optimizes the 2D and 3D networks, where the hybrid feature can be fully explored.
Figure~\ref{fig:result} presents some examples of liver and tumor segmentation results of our H-DenseUNet on the test dataset. We can observe that most small targets as well as large objects can be well segmented.
\subsection{Comparison with Other Methods on LiTS dataset}

There were more than 50 submissions in 2017 ISBI and MICCAI LiTS
challenges.
Both challenges employed the same training and test datasets for fair performance comparison.
Different from the ISBI challenge, more evaluation metrics have been added in the MICCAI challenge for comprehensive comparison.
The detailed results of top 15 teams on the leaderboard\revise{\footnote{https://competitions.codalab.org/competitions/17094\#results}}, including both ISBI and MICCAI challenges, are listed in Table~\ref{tab:result}. Our method (team name: xjqi, entry date: Nov. 17, 2017) outperformed other state-of-the-arts on the segmentation results of tumors and achieved very competitive performance for liver segmentation. 
For tumor burden evaluation, our method achieved the lowest estimation error and ranked the 1st place among all the teams. 
It is worth mentioning that we used ten entries on the test dataset for ablation analysis of our method. Since there is no validation set provided by challenge organizers,  the ablation experiments were performed on test dataset for fair comparison. Please note that the final result is just one of these entries, instead of multiple entries averages. 

Most of the top teams in the challenges employed deep learning based methods, demonstrating the effectiveness of CNN based methods in medical image analysis.
For example, \citet{han2017automatic,vorontsov2017liver} and \citet{bi2017automatic} all adopted 2D deep FCNs, where ResNet-like residual blocks were employed as the building blocks.
In addition, \citet{chlebus2017neural} trained the UNet architecture in two individual models, followed by a random forest classifier.
In comparison, our method with a 167-layer network consistently outperformed these methods, which highlighted the efficacy of 2D DenseUNet with pre-trained model.
Our proposed H-DenseUNet further advanced the segmentation accuracy for both liver and tumor, showing the effectiveness of the hybrid feature learning process.



Our method achieved the 1st place among all state-of-the-arts in the lesion segmentation and very competitive result to DeepX~\cite{yuan2017hierarchical} for liver segmentation.
Note that our method surpassed DeepX by a significant margin in the Dice per case evaluation for lesion, which is considered to be notoriously challenging and difficult.
Moreover, our result was produced by the single model while DeepX~\cite{yuan2017hierarchical} employed multi-model combination strategy to improve the results, showing the efficiency of our method in the clinical practice. 
\begin{table*}[!t]
	\centering \revise{
		\caption{Comparsion of tumor segmentation results on 3DIRCADb dataset.}
		\label{tab:3Ddataset_tumor} %
		{	\centering
			\begin{tabular}{c|c|c|c|c|c|c}
				\hline
				\multirow{2}{*}{Model}	& \multirow{2}{*}{Year}  &
				\multirow{2}{*}{VOE(\%)}	& \multirow{2}{*}{RVD(\%)}  &
				\multirow{2}{*}{ASD(mm)}	& \multirow{2}{*}{RMSD(mm)}  &
				\multirow{2}{*}{DICE}  \tabularnewline
				&  & & & & &       \tabularnewline
				\hline
				
				Unet ~\cite{chlebus2017neural} & 2017  & 62.55 $\pm$ 22.36  & 0.380 $\pm$ 1.95 & 11.11 $\pm$ 12.02  & 16.71 $\pm$ 13.81 & 0.51 $\pm$ 0.25  \tabularnewline
				\hline
				
				~\citet{christ2017automatic} & 2017 & - & - & - & - & 0.56 $\pm$ 0.26  \tabularnewline
				\hline
				ResNet  ~\cite{han2017automatic} & 2017  & 56.47 $\pm$ 13.62 & -0.41 $\pm$ 0.21  & 6.36 $\pm$ 3.77 & 11.69 $\pm$ 7.60 &  0.60 $\pm$ 0.12 \tabularnewline
				\hline
				ours &  & 49.72 $\pm$ 5.2 & -0.33 $\pm$ 0.10 & 5.293 $\pm$ 6.15 & 11.11 $\pm$ 29.14 & 0.65 $\pm$ 0.02  \tabularnewline
				\hline
				
				\hline
				\citet{foruzan2016improved}*  & 2016 & 30.61 $\pm$ 10.44 & 15.97 $\pm$ 12.04 & 4.18 $\pm$ 9.60 & 5.09 $\pm$ 10.71 & 0.82 $\pm$ 0.07  \tabularnewline
				\hline			
				\citet{wu20173d}* & 2017 & 29.04 $\pm$ 8.16 & -2.20 $\pm$ 15.88 & 0.72 $\pm$ 0.33 & 1.10 $\pm$ 0.49 & 0.83 $\pm$ 0.06   \tabularnewline
				\hline
				\citet{li2013likelihood}  $\dagger$ & 2013 & 14.4 $\pm$ 5.3 & -8.1 $\pm$ 2.1 & 2.4 $\pm$ 0.8 & 2.9 $\pm$ 0.7 & -  \tabularnewline
				\hline
				\citet{moghbel2016automatic}  $\dagger$ & 2016 & 22.78 $\pm$ 12.15 & 8.59 $\pm$ 18.78 & - & - & 0.75 $\pm$ 0.15   \tabularnewline
				\hline
				\citet{sun2017automatic}  $\dagger$ & 2017 & 15.6 $\pm$ 4.3 & 5.8 $\pm$ 3.5 & 2.0 $\pm$ 0.9 & 2.9 $\pm$ 1.5 & -  \tabularnewline
				\hline
				
				
				ours $\dagger $ & &  \textbf{11.68 $\pm$ 4.33} & \textbf{-0.01 $\pm$ 0.05} & \textbf{0.58 $\pm$ 0.46} & \textbf{1.87 $\pm$ 2.33} & \textbf{0.937 $\pm$ 0.02}  \tabularnewline
				\hline
			\end{tabular}
		}{	
		\begin{tablenotes}
			\centering
			\footnotesize
			\item Note: $\ast$ denotes the semi-automatic methods; $\dagger$ denotes the method use additional datasets;  - denotes the result is not reported.
		\end{tablenotes}
	}}
\end{table*}
\begin{table*}[!t]
	\centering
	\revise{
		\caption{Comparsion of liver segmentation results on 3DIRCADb dataset.}
		\label{tab:3Ddataset_liver} %
		{	\centering
			\begin{tabular}{c|c|c|c|c|c|c}
				\hline
				\multirow{2}{*}{Model}	& \multirow{2}{*}{Year}  &
				\multirow{2}{*}{VOE(\%)}	& \multirow{2}{*}{RVD(\%)}  &
				\multirow{2}{*}{ASD(mm)}	& \multirow{2}{*}{RMSD(mm)}  &
				\multirow{2}{*}{DICE}  \tabularnewline
				&  & & & & &       \tabularnewline
				\hline
				
				Unet ~\cite{chlebus2017neural} & 2017 & 14.21 $\pm$ 5.71 & -0.05 $\pm$ 0.10  & 4.33 $\pm$ 3.39 & 8.35 $\pm$ 7.54  & 0.923 $\pm$ 0.03 \tabularnewline
				\hline
				ResNet ~\cite{han2017automatic} & 2017 & 11.65 $\pm$ 4.06 & -0.03 $\pm$ 0.06 & 3.91 $\pm$ 3.95 &  8.11 $\pm$ 9.68 & 0.938 $\pm$ 0.02 \tabularnewline
				\hline
				~\citet{christ2017automatic} & 2017 & 10.7 & -1.4 & 1.5 & 24.0 & 0.943  \tabularnewline
				\hline
				ours &  & 10.02 $\pm$ 3.44 & -0.01 $\pm$ 0.05 & 4.06 $\pm$ 3.85 & 9.63 $\pm$ 10.41 & 0.947 $\pm$ 0.01  \tabularnewline
				\hline
				
				\hline
				\citet{li2015automaticliver} $\dagger$ & 2015 &  9.15 $\pm$ 1.44 & -0.07 $\pm$ 3.64 & 1.55 $\pm$ 0.39 & 3.15 $\pm$ 0.98 & -  \tabularnewline
				\hline
				\citet{moghbel2016automatic_liver}$\dagger$ & 2016& 5.95 & 7.49 & - & -  &  
				0.911
				\tabularnewline
				\hline
				\citet{lu2017automatic} $\dagger$ & 2017 &  9.36 $\pm$ 3.34 & 0.97 $\pm$ 3.26 & 1.89 $\pm$ 1.08 & 4.15 $\pm$ 3.16 & -  \tabularnewline
				\hline
				ours $\dagger$ & &  \textbf{3.57 $\pm$ 1.66} & \textbf{0.01 $\pm$ 0.02} & \textbf{1.28 $\pm$ 2.02} & \textbf{3.58 $\pm$ 6.58} & \textbf{0.982 $\pm$ 0.01}  \tabularnewline
				\hline
				
			\end{tabular}
		}{	
		\begin{tablenotes}
			\centering
			\small
			\item Note: $\dagger$ denotes the method use additional datasets. - denotes the result is not reported.
		\end{tablenotes}
	}}
\end{table*}

\subsection{\revise{Comparison with Other Methods on 3DIRCADb Dataset}}
To validate the effectiveness and robustness of our method, we also conduct experiments on 3DIRCADb dataset~\cite{soler20103d}, which is publicly available and offers a higher variety and complexity of livers and lesions.
\revise{
	Table~\ref{tab:3Ddataset_tumor} and Table~\ref{tab:3Ddataset_liver} show the comparison of the tumor and liver segmentation performance on the 3DIRCADb dataset.
	We compared our method with the state-of-the-art method~\cite{christ2017automatic} on the 3DIRCADb dataset by running experiments through cross-validation, as the way used in~\cite{christ2017automatic}.  
	We can see that our method achieved the better performance than~\cite{christ2017automatic} on both lesion and liver segmentation accuracy, with 9.0\% and 0.4\% improvement on DICE, respectively. 
	To further validate the effectiveness of our method, we ran experiments with methods of Unet~\cite{chlebus2017neural} and ResNet architecture~\cite{han2017automatic} respectively, where the training setting keeps the same with~\citet{{christ2017automatic}}. 
	From Table~\ref{tab:3Ddataset_tumor} and Table~\ref{tab:3Ddataset_liver}, we can see that our method still outperforms Unet~\cite{chlebus2017neural} and ResNet~\cite{han2017automatic} on the 3DIRCADb dataset, with 14.0\% and 5.0\% improvement on DICE for tumor segmentation respectively. 
	The experimental comparison validated the superiority of our proposed method in comparison with other methods.}

\begin{figure}[!t]
	\centering
	\includegraphics[width=0.98\linewidth]{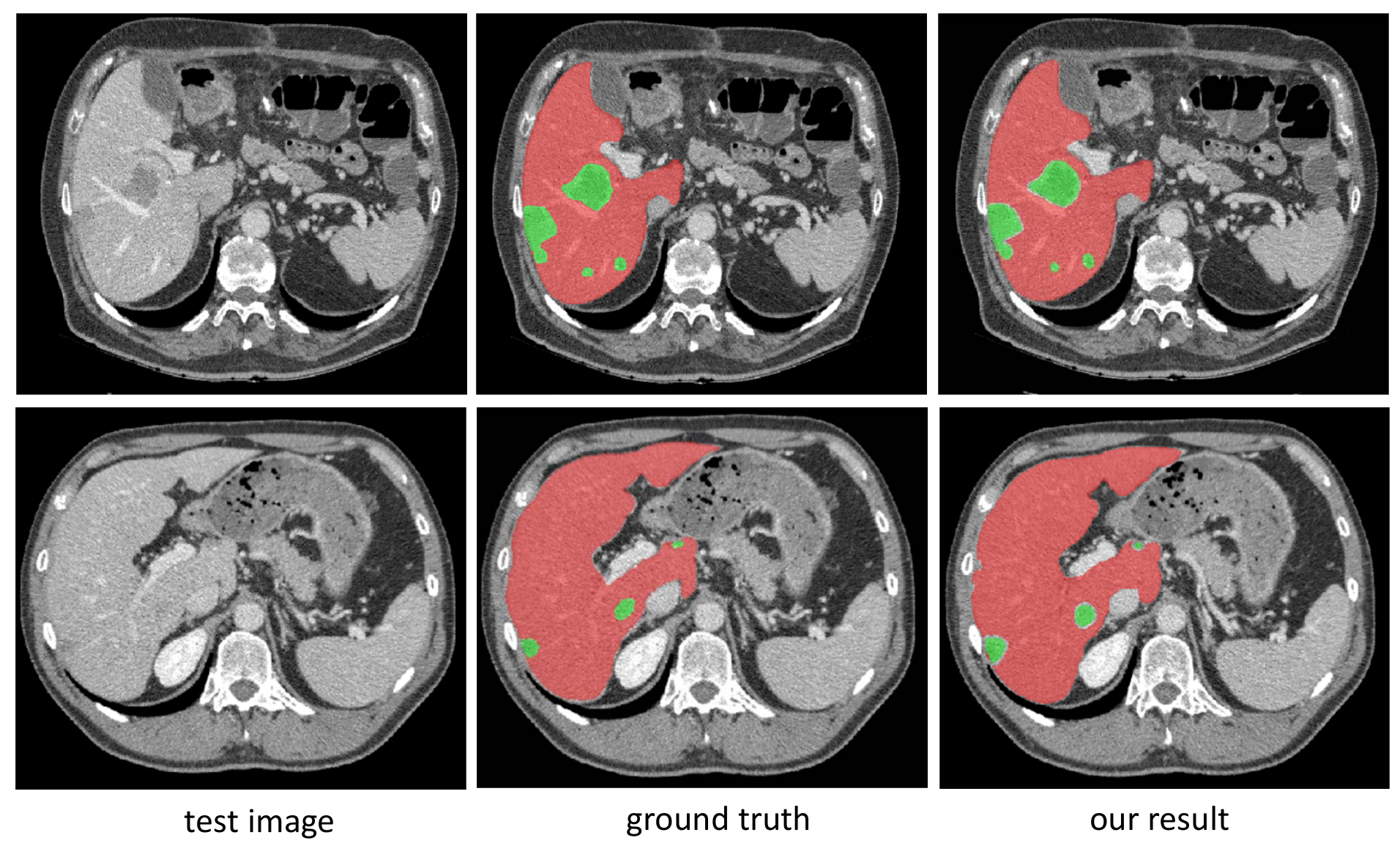}
	\caption{\revise{Examples of our segmentation results on the 3DIRCADb dataset.}}
	\label{fig:3d}\centering
\end{figure}

\revise{To have a comprehensive comparison with liver tumor segmentation methods, we listed the reported tumor and liver segmentation results on the 3DIRCADb dataset below the bold line in Table~\ref{tab:3Ddataset_tumor} and Table~\ref{tab:3Ddataset_liver}, respectively. 
	Note that except experiments~\cite{chlebus2017neural} and \cite{han2017automatic}, all other experiment results are the reported values in the original papers.
	It is worth noting that most liver tumor segmentation methods~\cite{sun2017automatic,lu2017automatic,li2013likelihood,moghbel2016automatic,li2015automaticliver,moghbel2016automatic_liver} 	
	utilized additional datasets for training and tested on the 3DIRCADb dataset. 
	For example, ~\citet{li2013likelihood}, ~\citet{sun2017automatic} and ~\citet{lu2017automatic} collected additional clinical data from hospitals as the training set.
	~\citet{moghbel2016automatic} utilized additional the MIDAS dataset while~\citet{li2015automaticliver} used the SLIVER07 dataset in the training, respectively. 
	In addition, ~\citet{foruzan2016improved} and~\citet{wu20173d} achieved good results on tumor segmentation by semi-automatic methods.
	Actually, these methods cannot be compared directly with each other due to the differences in the training dataset and whether is fully-automatic or not.
	However, to some extent, the reported results on the 3DIRCADb dataset can reflect the state-of-the-art performance for the lesion and liver segmentation task. 
	Here, we employed the LiTS dataset as the additional dataset. 
	Specifically, we directly tested the well-trained model from 2017 LiTS dataset on the 3DIRCADb dataset. 
	As shown in Table~\ref{tab:3Ddataset_tumor} and Table~\ref{tab:3Ddataset_liver}, our method achieves the best tumor and liver segmentation results on the 3DIRCADb dataset, surpassing the state-of-the-art result largely, with 10.7\% and 7.1\% improvement on DICE for tumor and liver segmentation respectively. 
	The promising result indicates the effectiveness and good generalization capability of our method.
	On the other hand, such a good result is also attributed to the LiTS dataset, which contains a huge amount of training data with large variations, and the ability of our method to extract discriminative features from this dataset.
	Figure~\ref{fig:3d} shows some examples of the results on the 3DIRCADb dataset. 
	It is obvious that our method can well segment the liver and liver lesions from challenging raw CT scans.}

\setlength{\textfloatsep}{3pt plus 1.0pt minus 2.0pt}

\section{Discussion}
Automatic liver and tumor segmentation plays an important role in clinical diagnosis. It provides the precise contour of the liver and any tumors inside the anatomical segments of the liver, which assists doctors in the diagnosis process. 
In this paper, we present an end-to-end training system to explore hybrid features for automatic liver lesion segmentation, where the 3D contexts are effectively probed under the auto-context mechanism.
Through the hybrid fusion learning of intra-slice and inter-slice features, the segmentation performance for liver lesion has been improved, which demonstrates the effectiveness of our H-DenseUNet.
Moreover, compared with other 3D networks~\cite{cciccek20163d,dou20163d}, our method probes 3D contexts efficiently. This is crucial in the clinical practice, especially when huge amount of 3D images, containing large image size and a number of slices, are increasingly accumulated in the clinical sites.

\revise{To show the generalization capability of our method in the clinical practice, we tested our trained model from the LiTS dataset on the 3DIRCADb dataset, and it achieved the state-of-the-art results on both liver and tumor segmentation, with 98.2\% and 93.7\% on DICE. 
	The promising results achieved on the 3DIRACDb dataset also validated that our method is not simple overtraining, but actually is effective to generalize to different dataset under different data collection conditions.}

\begin{figure}[!t]
	\centering
	\includegraphics[width=1.0\linewidth]{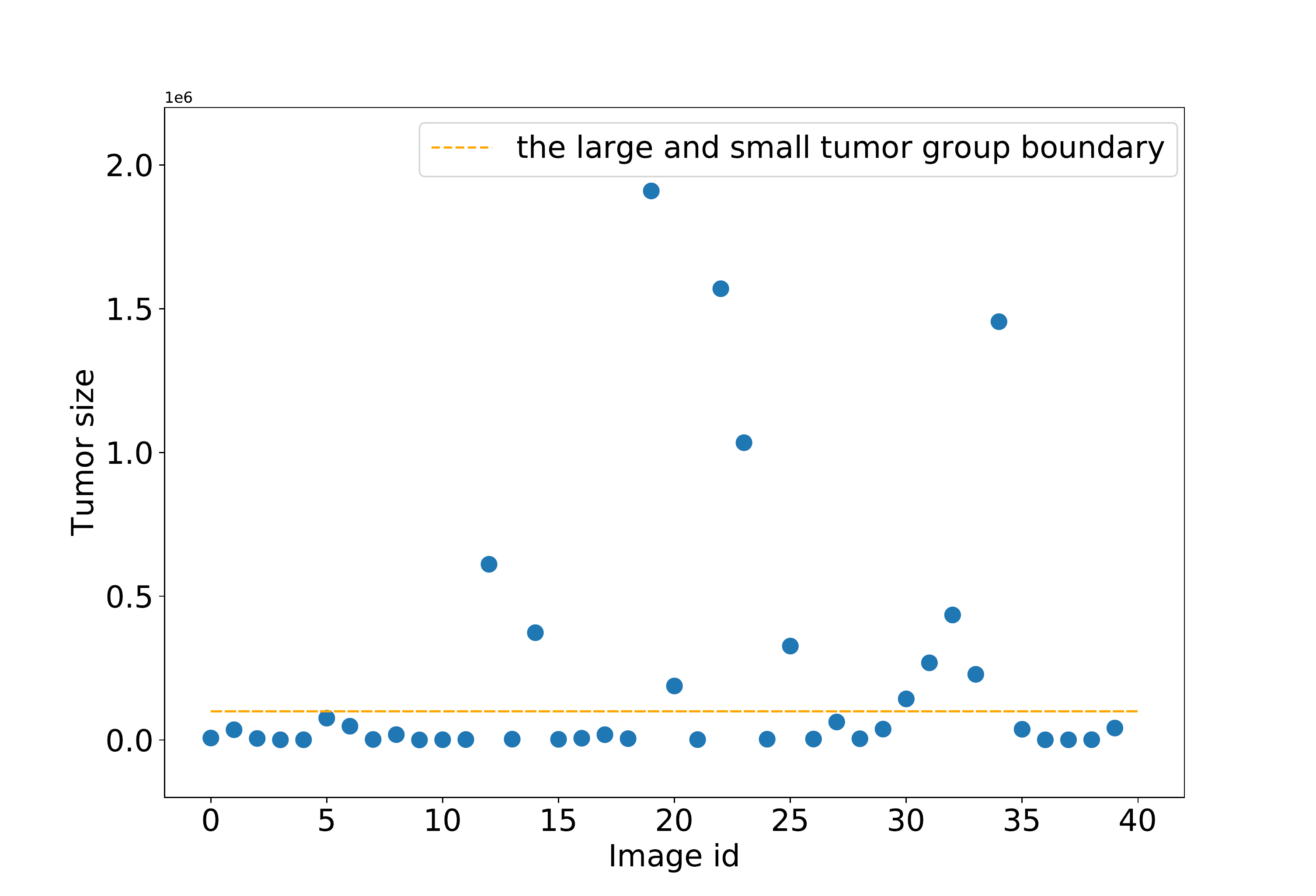}
	\caption{Tumor size (tumor voxels number) in each patient of our validation dataset. We define the orange line to seperate the large-tumor and the small-tumor group.}
	\label{fig:voxels}\centering
\end{figure}

To have a better understanding about the performance gains, we analyze the effectiveness of our method regarding the liver tumor size in each patient.
Figure~\ref{fig:voxels} shows the tumor size value of 40 CT volume data in our validation dataset, where the tumor size is obtained by summing up tumor voxels in each ground-truth image. It is observed that the dataset has large variations of the tumor size.
For comparison, we divide the dataset into the large-tumor group and the small-tumor group by the orange line in Figure~\ref{fig:voxels}.
From Table~\ref{tab: tumor size performance}, we can observe that our method improves the segmentation accuracy by 1.48 (Dice:\%) in the whole validation dataset. We can also observe that the large-tumor group achieves 2.35 (Dice:\%) accuracy improvements while the score for the small-tumor group is slightly advanced, with 1.1 (Dice:\%).
From the comparison, we claim that the performance gain is mainly attributed to the improvement of the large-tumor data segmentation results.
This is mainly because that the H-DenseUNet mimics the diagnosis process of radiologists, where tumors are delineated by observing several adjacent slices, especially for tumors have blurred boundaries. Once the blurred boundaries are well segmented, the segmentation accuracy for the large-tumor data can be improved by a large margin.
Although the hybrid feature still contributes to the segmentation of small tumors, the improvement is limited since small tumors usually occur in fewer slices.
In the future, we will focus on the segmentation for small liver tumors. Several potential directions will be taken into considerations for tackling small liver tumor problem, i.e., multi-scale representation structure~\cite{kamnitsas2017efficient} and deep supervision~\cite{dou20163d}.~\revise{Recently, perceptual generative adversarial networks (GANs) have been proposed for small object detection and classification~\cite{li2017perceptual,frid2018gan}. For example, ~\citet{li2017perceptual} generated superresolved representations for small objects by discovering the intrinsic structural
	correlations between small-scale and large-scale objects, which may also be a potential direction for handling this challenging problem.}
\begin{table}[!t]
	\centering
	\caption{Effectiveness of our method regarding to the tumor size (Dice: \%).}
	\label{tab: tumor size performance} %
	{	
		\begin{tabular}{c|c|c|c}
			\hline
			& Total & Large-tumor group & Small-tumor group    \tabularnewline
			\hline
			Baseline & 43.56 &  58.24  & 41.08       \tabularnewline
			\hline
			H-DenseUNet  &  45.04 (+1.48) & 60.59 (+2.35)  &  42.18  (+1.1) \tabularnewline
			\hline
		\end{tabular}
		{	
			\begin{tablenotes}
				\centering
				\small
				\item Note: Baseline is the 2D DenseUNet with pre-trained model.
			\end{tablenotes}
		}
	}
\end{table}

Another key that should be explored in the future study is the potential depth for the H-DenseUNet. In our experiments, we trained the network using data parallel training, which is an effective technique to speed up the gradient descent by paralleling the computation of the gradient for a mini-batch across mini-batch elements. However, the model complexity is restricted by the GPU memory. 
In the future, to exploit the potential depth of the H-DenseUNet, we can train the network using model parallel training, where different portions of the model computation are done on distributed computing infrastructures for the same batch of examples.
This strategy maybe another possible direction to further improve the liver tumor segmentation performance.

\section{Conclusion}
We present an end-to-end training system H-DenseUNet for liver and tumor segmentation from CT volumes, which is a new paradigm to effectively probe high-level representative intra-slice and inter-slice features, followed by optimizing the features through the hybrid feature fusion layer. 
The architecture gracefully addressed the problems that 2D convolutions ignore the volumetric contexts and 3D convolutions suffer from heavy computational cost. 
Extensive experiments on the dataset of 2017 LiTS and 3DIRCADb dataset demonstrated the superiority of our proposed H-DenseUNet. 
With a single-model basis, our method excelled others by a large margin on lesion segmentation and achieved very competitive result on liver segmentation on the LiTS Leaderboard.


\section{Acknowledgments}
This work was supported by the grants from the Research Grants Council of the Hong Kong Special Administrative Region (Project Nos.  GRF 14202514 and GRF 14203115).

\bibliographystyle{IEEEtranN}
\bibliography{refs}
\ifCLASSOPTIONcaptionsoff
  \newpage
\fi
\end{document}